\pdfoutput=1 

\documentclass[journal]{IEEEtran}
\ifCLASSINFOpdf
   \usepackage{graphicx}
   \usepackage{mathtools}
\usepackage{multirow}
\usepackage{xcolor}
\usepackage{amsmath}
\usepackage[pagebackref=true,breaklinks=true,letterpaper=true,colorlinks,bookmarks=false]{hyperref}
\usepackage{amsfonts}
\usepackage{makecell}

\else
\fi
\begin{document}
%
\title{3D Shape Reconstruction from Free-Hand Sketches}
%
%
%

\author{Jiayun Wang, Jierui Lin, Qian Yu, Runtao Liu, Yubei Chen, and Stella X. Yu
\thanks{All the authors except Q. Yu are with UC Berkeley / ICSI, Berkeley, CA, USA (e-mail: \{peterwg,jerrylin0928,runtao\_liu,yubeic,stellayu\}@berkeley.edu).}
\thanks{Q. Yu is with Beihang University, Beijing, China (e-mail: qianyu@buaa.edu.cn) and UC Berkeley / ICSI, Berkeley, CA, USA.}
}

%
%

\markboth{}
{Wang \MakeLowercase{\textit{et al.}}: 3D Shape Reconstruction from Free-Hand Sketches}
%



\maketitle

%

\IEEEpeerreviewmaketitle

\begin{abstract}

Sketches are the most abstract 2D representations of real-world objects. Although a sketch usually has geometrical distortion and lacks visual cues, humans can effortlessly envision a 3D object from it. This suggests that sketches encode the information necessary for reconstructing 3D shapes. Despite great progress achieved in 3D reconstruction from distortion-free line drawings, such as CAD and edge maps, little effort has been made to reconstruct 3D shapes from free-hand sketches. 
We study this task and aim to enhance the power of sketches in 3D-related applications such as interactive design and VR/AR games. 

Unlike previous works, which mostly study distortion-free line drawings, our 3D shape reconstruction is based on free-hand sketches. A major challenge for free-hand sketch 3D reconstruction comes from the insufficient training data and free-hand sketch diversity, e.g. individualized sketching styles.
We thus propose data generation and standardization mechanisms.
Instead of distortion-free line drawings, synthesized sketches are adopted as input training data. Additionally, we propose a sketch standardization module to handle different sketch distortions and styles. Extensive experiments demonstrate the effectiveness of our model and its strong generalizability to various free-hand sketches. Our \href{https://github.com/samaonline/3D-Shape-Reconstruction-from-Free-Hand-Sketches}{code} is publicly available.
\end{abstract}


\begin{figure*}[t!]
\centering
  \includegraphics[width=1\textwidth]{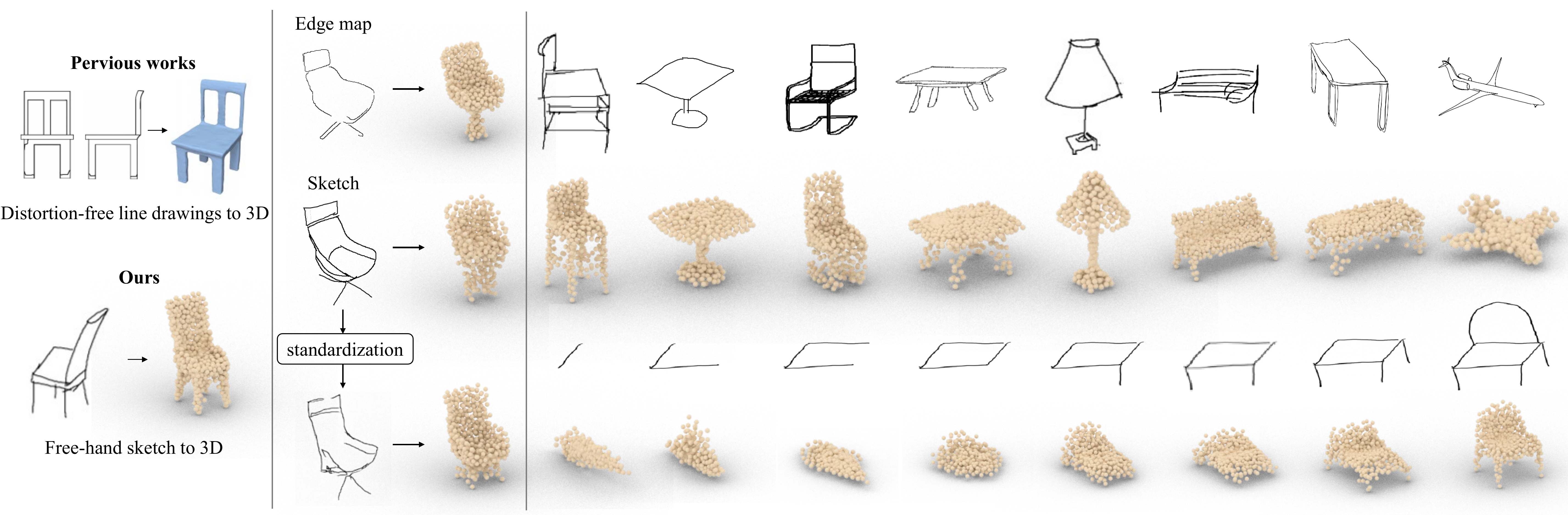}
  \caption{ 
  We study 3D reconstruction from a single-view free-hand sketch, differing from previous works \cite{delanoy20183d, wang2015sketch, lun20173d} which use multi-view distortion-free line-drawings as training data. ({\bf Left}).
 While previous works \cite{delanoy20183d, wang2015sketch} employ distortion-free line drawings (e.g. edge-maps) as proxies for sketches, we show that a model trained on synthesized sketches can generalize better to free-hand sketches. We also show that the proposed sketch standardization module makes the method generalizes well to free-hand sketches by dealing with different sketching styles as well as distortions ({\bf Middle}). Our model reconstructs 3D shapes from various free-hand sketches and may unleash many practical applications such as real-time 3D modeling with sketches  ({\bf Right}). \textcolor{black}{A real-time sketching demo is available at \href{https://streamable.com/z76m47}{https://streamable.com/z76m47}.}}
   \label{fig:intro}
\end{figure*}

\section{Introduction}

\IEEEPARstart{H}{uman} free-hand sketches are the most abstract 2D representations for 3D visual perception. Although a sketch may consist of only a few colorless strokes and exhibit various deformation and abstractions, humans can effortlessly envision the corresponding real-world 3D object from it. 
It is of interest to develop a computer vision model that can replicate this ability.
Although sketches and 3D representations have drawn great interest from researchers in recent years, these two modalities have been studied relatively independently. We explore the plausibility of bridging the gap between sketches and 3D, and build a computer vision model to recover 3D shapes from sketches. Such a model will unleash many applications, like interactive CAD design and VR/AR games.

With the development of new devices and sensors, sketches and 3D shapes, as representations of real-world objects beyond natural images, become increasingly important. The popularity of touch-screen devices makes sketching not a privilege to professionals anymore and increasingly popular. Researchers have applied sketch in tasks like image retrieval \cite{yu2016sketch,sangkloy2016sketchy} and image synthesis \cite{liu2019unpaired,ghosh2019interactive} to leverage its power in expression. Furthermore, as depth sensors, such as structured light device, LiDAR, and TOF cameras, become more ubiquitous, 3D data become an emerging modality in computer vision. 3D reconstruction, the process of capturing the shape and appearance of real objects, is an essential topic in 3D computer vision. 3D reconstruction from multi-view images has been studied for many years \cite{fuentes2015visual, choy20163d, ozyecsil2017survey}. Recent works \cite{richter2015discriminative, dibra2017human, DBLP:journals/corr/FanSG16} have further explored 3D reconstruction from a single image.

Despite these trends and progress, there are limited works connecting 3D and sketches. We argue that sketches are abstract 2D representations of 3D perception, and it is of great significance to study sketches in a 3D-aware perspective and build connections between two modalities. Researchers have explored the potential of \textit{distortion-free} line drawings (e.g. edge maps) for 3D modeling \cite{li2017bendsketch,Xu:2014:True2Form,li2018robust}. These works are based on \textit{distortion-free} line drawings and generalize poorly to free-hand sketches (Fig.\ref{fig:intro}({Left})). Furthermore, the role of line drawings in such works is to provide geometrical information for the subsequent 3D modeling. Some other works \cite{delanoy20183d,lun20173d} employ neural networks to reconstruct 3D shapes directly from line drawings. However, their decent reconstructions come with two major limitations: 
\textbf{1)} they use distortion-free line drawings as training data, which makes such models hard to generalize to free-hand sketches; 
\textbf{2)} they usually require inputs depicting the object from multi-views to achieve satisfactory outcomes. 
Therefore, such methods cannot reconstruct the 3D shape from a single-view free-hand sketch well, as we show later in the experiment section. 
Other works such as \cite{wang2015sketch,he2018triplet} tackle 3D retrieval instead of 3D shape reconstruction from sketches. Retrieved shapes come from the pre-collected gallery set and may not resembles novel sketches well.
Overall, reconstructing a 3D shape from a single \textit{free-hand} sketch is still left not well explored.

In this work, we explore single-view free-hand sketch-based 3D reconstruction (Fig.\ref{fig:intro}({Middle})). 
A free-hand sketch is defined as a line drawing created without any additional  tool. As an abstract and concise representation, it is different from distortion-free line drawings (e.g. edge maps) since it commonly has some spatial distortions, but it can still reflect the essential geometric shape.
3D reconstruction from sketch is challenging due to the following reasons: 
\begin{enumerate}
    \item Data insufficiency. Paired sketch-3D datasets are rare although there exist several large-scale sketch datasets and 3D shape datasets, respectively. Furthermore, collecting sketch-3D pairs can be very time-consuming and expensive than collecting sketch-image pairs, as each 3D shape could be sketched from various viewing angles. 

    \item There is a misalignment between two representations. A sketch depicts an object from a certain view while a 3D shape can be viewed from multiple angles due to the encoded depth information. 
    
    \item Due to the nature of hand drawing, a sketch is usually geometrically imprecise with a individual style compared to the real object. Thus a sketch can only provide suggestive shape and structural information. In contrast, a 3D shape is faithful to its corresponding real-world object with no geometric deformation.
\end{enumerate}

To address these challenges, we propose a single-view sketch-to-3D shape reconstruction framework. Specifically, it takes a sketch from an \textit{arbitrary} angle as input and reconstructs a 3D point cloud. Our model cascades a sketch standardization module $U$ and a reconstruction module $G$. $U$ handles various sketching styles/distortions and transfers inputs to standardized sketches  while $G$ takes a standardized sketch to reconstruct the 3D shape (point cloud) \textit{regardless of the object category}.
The key novelty lies in the mechanisms we propose to tackle the data insufficiency issue.
Specifically, we first train an photo-to-sketch model on unpaired large-scale datasets.
Based on the model, sketch-3D pairs can be automatically generated from 2D renderings of 3D shapes. Together with the  standardization module $U$ which unifies input sketch styles, the synthesized sketches provide sufficient information to train the reconstruction model $G$. We conduct extensive experiments on a composed sketch-3D dataset, spanning 13 classes, where sketches are synthesized and 3D objects come from the ShapeNet dataset \cite{shapenet2015}. Furthermore, we collect an evaluation set, which consists of 390 real sketch-3D pairs. \textcolor{black}{ Results demonstrate that our model can reconstruct 3D shapes with certain geometric details from real sketches under different styles, stroke line-widths, and object categories.} Our model also enables practical applications such as real-time 3D modeling with sketches (Fig.\ref{fig:intro}({Right})).

To summarize, our work makes the following contributions:
\begin{enumerate}
    \item We pioneer to study the plausibility of reconstructing 3D shapes from single-view free-hand sketches. The proposed model demonstrates its robust performance on real sketches.
    
    \item We propose a novel framework for this task and explore various design choices to overcome challenges.
    
    \item To handle the data insufficiency problem, we propose to train on synthetic sketches generated by a GAN-based model. Moreover, a sketch standardization module is introduced to make the model generalize to free-hand sketches better. The proposed sketch generation method can also ease the generalization issue for many other stroke-based applications, such as OCR.
\end{enumerate}

\begin{figure*}[t!]
\centering
  \includegraphics[width=1\textwidth]{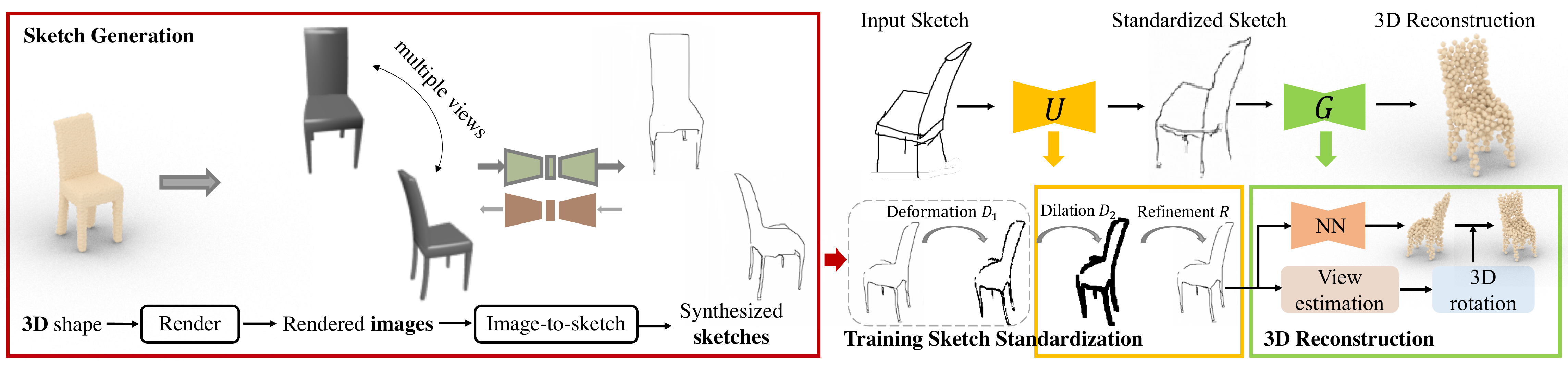}
  \caption{ The pipeline of our model. The model consists of three major components: sketch generation (\textcolor{black}{red box}), sketch  standardization (\textcolor{orange}{orange box}), and 3D reconstruction (\textcolor{green}{green box}). To generate synthesized sketches for training the model, we first render 2D images for a 3D shape (in the format of mesh) from multiple viewpoints, and then employ an image-to-sketch translation model to generate sketches of corresponding views. The standardization module is introduced to handle sketches with different styles and distortions. \textit{Deformation $D_2$} is only used in training for augmentation such that the model is robust to geometric distortions of sketches. During evaluation, free-hand sketches are dilated ($D_1$) and refined ($R$) so their style matches that of training sketches. In the 3D reconstruction network, a view estimation module is adopted to align the output's view and the corresponding ground-truth 3D shape. }
  \label{fig:network}
\end{figure*}

\section{Related Works}

\subsection{3D Reconstruction from Images} 
SfM \cite{ozyecsil2017survey} and SLAM \cite{ fuentes2015visual} achieve success in handling multi-view 3D reconstructions in various real-world scenarios. However, their reconstructions can be limited by insufficient input viewpoints and 3D scanning data. Deep-learning-based methods have been proposed to further improve reconstructions by completing 3D shapes with occluded or hollowed-out areas \cite{yang2018dense, choy20163d, kar2017learning}.

In general, recovering the 3D shape from a single-view image is an ill-posed problem. Attempts to tackle the problem include 3D shape reconstructions from silhouettes \cite{dibra2017human}, shading \cite{richter2015discriminative}, and texture \cite{witkin1981recovering}. However, these methods need strong presumptions and expertise in natural images \cite{zhang2019realpoint3d}, limiting their usage in real-world scenarios. Generative adversarial networks (GANs) \cite{goodfellow2014generative} and variational autoencoders (VAEs) \cite{kingma2013auto} have achieved success in image synthesis and enabled \cite{wu2016learning} 3D shape reconstruction from a single-view image. Fan \textit{et al.} \cite{DBLP:journals/corr/FanSG16} further adopt point clouds as 3D representation, enabling models to reconstruct certain geometric details from an image. They may not directly work on sketches as many visual cues are missing.

3D reconstruction networks are designed differently depending on the output 3D representation. 3D voxel reconstruction networks \cite{tatarchenko2017octree, hane2017hierarchical, xie2019pix2vox} benefit from many image processing networks as convolutions are appropriate for voxels. They are usually constrained to low resolution due to the computational overhead. Mesh reconstruction networks \cite{wang2018pixel2mesh, kolotouros2019convolutional} are able to directly learn from meshes, where they suffer from topology issues and heavy computation \cite{pan2019deep}. We adopt point cloud representation as it can capture certain 3D geometric details with low computational overhead. Reconstructing 3D point clouds from images has been shown to benefit from well-designed network architectures \cite{DBLP:journals/corr/FanSG16, mandikal2019dense}, latent embedding matching \cite{mandikal20183d}, additional image supervision \cite{mandikal2018capnet}, etc.

\subsection{Sketch-Based 3D Retrievals and Reconstructions} 
Free-hand sketches are used for 3D shape retrieval \cite{wang2015sketch,he2018triplet} given their power in expression. However, retrieval methods are significantly constrained by the gallery dataset. Precise sketching is also studied in the computer graphics community for 3D shape modeling or procedural modeling \cite{li2017bendsketch,huang2016shape,li2018robust}. These works are designed for professionals and require additional information for shape modeling, e.g., surface-normal, procedural model parameters. 

Delanoy \textit{et al.} \cite{delanoy20183d} first employ neural networks to learn 3D voxels from line-drawings. While it achieves impressive performance, this model has several limitations: 1) The model uses distortion-free edge map as training data. While working on some sketches with small distortions, it cannot generalize to general free-hand sketches. 2) The model requires multiple inputs from different viewpoints for a satisfactory result.  These limitations prevent the model from generalizing to real free-hand sketches.
\textcolor{black}{ Recent works also explore reconstructing 3D models from sketches with direct shape optimization \cite{han2020reconstructing},  differential renderer \cite{zhang2021sketch2model}, and  unsupervised learning \cite{wang2018unsupervised}. }
Unlike the existing works, the proposed method in this work reconstructs the 3D point cloud based on a single-view free-hand sketch. Our model may make 3D reconstruction and its applications more accessible to the general public.

\begin{figure*}[t!]
\centering
  \includegraphics[width=1\textwidth]{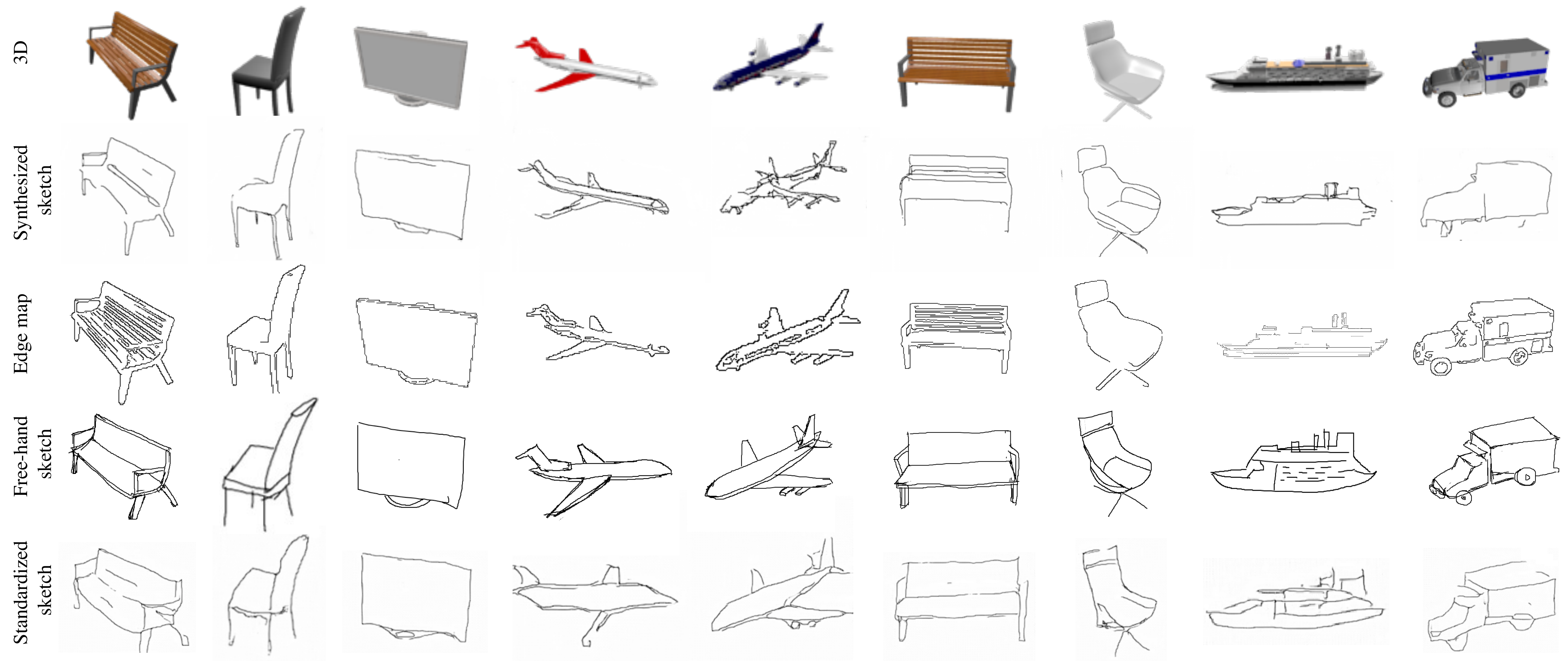}
  \caption{ 
  \textcolor{black}{Synthesized sketches (2nd row) are visually more similar to free-hand sketches (4th row) than edge maps (3rd row) as they contain distortions and emphasize perceptually significant contours. Note that newly collected free-hand sketches (4th row) are used for evaluation only. Additionally, applying the standardization module, the standardized free-hand sketches (last row) share a uniform style similar to that of training data. }
  }
  \label{fig:sketch_gen}
\end{figure*}

\section{3D Reconstruction from Sketches}
The proposed framework is composed of three modules (Fig.\ref{fig:network}). To deal with the data insufficiency issue, we first synthesize sketches as the training set (Sec.~\ref{sec:skgen}). The module $U$ transfers an input sketch to a standardized sketch (Sec.~\ref{sec:skuni}). After that, the module $G$ takes the standardized sketch to reconstruct a 3D shape (point clouds) (Sec.~\ref{sec:3Drec}). In Sec.~\ref{sec:dataset}, we present details of a new sketch-3D dataset, which is collected for evaluating the proposed model. 
\subsection{Synthetic Sketch Generation}
\label{sec:skgen}

\label{sec:sk_gen}
To the best of our knowledge, there exists no paired sketch-3D dataset. While it is possible to resort to edge maps \cite{delanoy20183d}, edge maps are different from sketches (as shown in the 3rd and 4th rows of Fig.\ref{fig:sketch_gen}). We show that the reconstruction model trained on edge maps cannot generalize well to real free-hand sketches  in Sec.~\ref{sec:edgemap}. Thus it is crucial to find an efficient and reliable way to synthesize sketches for 3D shapes. Inspired by \cite{liu2019unpaired}, we employ a generative model to synthesize sketches from rendered images of 3D shapes. 
Fig.\ref{fig:network}(Left) depicts the procedure. Specifically, we first render $m$ images for each 3D shape, where each image corresponds to a particular view of a 3D shape. \textcolor{black}{ We then adopt the model introduced in \cite{liu2019unpaired} to synthesize gray-scale sketches images, denoted as $\{S_i|S_i\in \mathbb{R}^{W\times H}\}$, as our training data. $W, H$ refer to the width and height of a sketch image.}

{\color{black}\subsection{Sketch  Standardization}
\label{sec:skuni}

Free-hand sketches usually have strong individual styles and geometric distortions.
Due to the large gap between the free-hand sketches and the synthesized sketches, directly using the synthesized sketches as training data would not lead to a robust model. The main issues are that the synthesized sketches have a uniform style and they do not contain enough geometric distortions. Rather, the synthesized sketches can be treated as an intermediate representation if we can find a way to project a free-hand sketch to the synthesized sketch domain. We propose a zero-shot domain translation technique, the sketch standardization module, to achieve this domain adaption goal without using the free-hand sketches as the training data. The training of the sketch standardization module only involves synthesized sketches. The general idea is to project a distorted synthesized sketch to the original synthesized sketch. The training consists of two parts: \textbf{1) } since the free-hand sketches usually have geometric distortions, we apply predefined distortion augmentation to the input synthesized sketches first. \textbf{2) } A geometrically distorted synthesized sketch still has a different style and line style compared to the free-hand sketches. Thus, the first stage of the standardization is to apply a dilation operation. The dilation operation would project distorted synthesized sketches and the free-hand sketches to the same domain. Then, a refinement network follows to project the dilated sketch back to the synthesized sketch domain. 

In summary, as shown in Fig.\ref{fig:network}, the standardization module $U$ first applies dilation operator $D_2$ to the input sketch, which is followed by a refinement operator $R$ to transfer to the standardized synthesized-sketch style (or training-sketch style) $\widetilde{S}_i$, i.e. $U=R\circ D_2$. $R$ is implemented as an image translation network. During training, a synthesized sketch $\widetilde{S}_i$ is first augmented by the deformation operator $D_1$ to mimic the drawing distortion, and then $U$ aims to project it back to $\widetilde{S}_i$. Please note that $D_1$ would not be used during the testing.
We illustrate the standardization process in Fig.\ref{fig:network}(Right) with more details in the following.

\textbf{Deformation.} During training of $U$, each synthesized input sketch is randomly deformed with moving least squares \cite{schaefer2006image} to make the distortion is local and rigid. Specifically, we randomly sample a set of control points on sketch strokes and denote them as $p$, and denote the corresponding deformed point set as $q$. Following moving least squares, we solve for the best affine transformation $l_v(x)$ such that:
\begin{align}
\label{eq:rigid}
    \min \sum_i w_i | l_v(p_i) -q_i|^2
\end{align}
where $p_i$ and $q_i$ are row vectors and weights $w_i = \frac{1}{|p_i - v|^{2 \alpha}}$. 
Affine transformation can be written as $l_v(p_i) = p_i M+T$. We add constraint $M^TM = I$ to make the deformation is rigid to avoid too much distortion.
Details of solving Eqn.\ref{eq:rigid} can be found in \cite{schaefer2006image}.

\textbf{Style Translation.}
Adapting to unknown input free-hand sketch style during inference can be considered as a zero-shot domain translation problem, which is challenging. 
Inspired by \cite{yang2020deep}, we first dilate the augmented training sketch strokes with 4 pixels and then use image-to-image translation network Pix2Pix \cite{isola2017image} to translate the dilated sketches to the un-distorted synthesized sketches.
During inference, we also dilate the free-hand sketches and apply the trained Pix2Pix model such that the style of an input free-hand sketch could be adapted to the synthesized sketch style during training. The dilation step can be considered as introducing uncertainty for the style adaption.
Further, we show in Section \ref{sec:ssmexp} that the proposed style standardization module could be used as a general zero-shot domain translation technique, which generalizes to more applications such as sketch classification and zero-shot image-to-image translation.

\textbf{A More General Message: Zero-Shot Domain Translation.} We illustrate in Fig.\ref{fig:zerodemo} a more general message of the standardization module: it can be considered as a general method for zero-shot domain translation. Consider the following problem: we would like to build a model to transfer domain $X$ to domain $Z$ but we do not have any training data from domain $X$. We propose a general idea to solve this problem is to build an intermediate domain $Y$ as a bridge such that: 1) we can translate data from domain $X$ to domain $Y$ and 2) we can further translate data from domain $Y$ to domain $Z$. We give two examples in the caption of Fig.\ref{fig:zerodemo} and provided experimental results in Section \ref{sec:ssmexp}.

\begin{figure}[t!]
\centering
  \includegraphics[width=1\columnwidth]{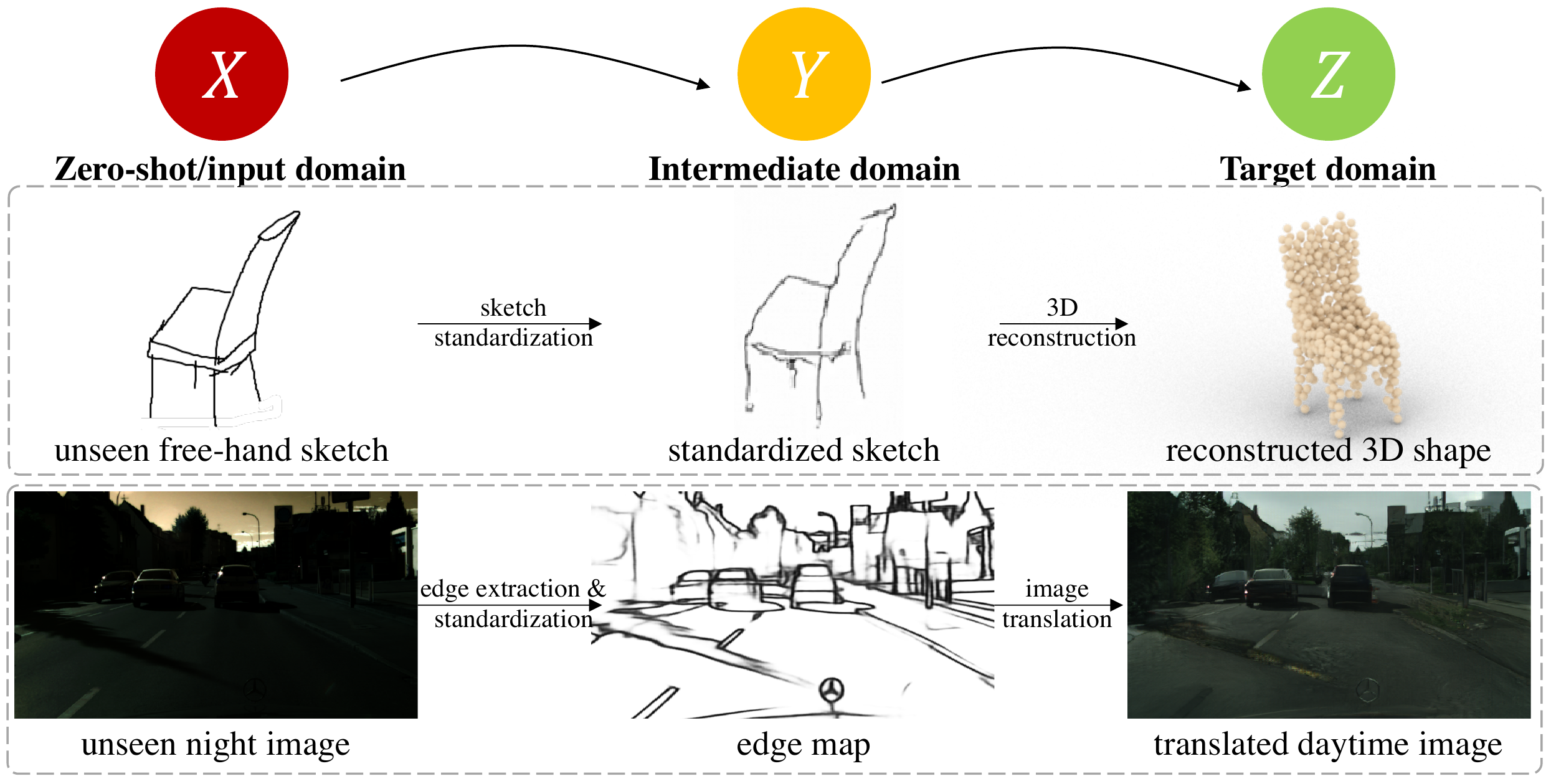}
  \caption{ \textcolor{black}{Sketch standardization module can be considered as a general zero-shot domain translation method. Given a sample from a zero-shot (input) domain $X$, we first translate it to a universal intermediate domain $Y$, and finally to the target domain $Z$. In the first example (second row), the input domain is an unseen free-hand sketch. With sketch standardization, it is translated to an intermediate domain: standardized sketch, which shares similar style as synthesized sketch for training. With 3D reconstruction, the standardized sketch can be translated to the target domain: 3D point clouds. In the second example (last row), the input domain is an unseen nighttime image. With edge extraction, it gets translated to an intermediate domain: edge map. With the image-to-image translation model, the standardized edge map can be translated to the target domain: daytime image. } }
  \label{fig:zerodemo}
\end{figure}

}

\subsection{Sketch-Based 3D Reconstruction}
\label{sec:3Drec}
Our 3D reconstruction network $G$ (pipeline in Fig.\ref{fig:network}(Right)) consists of several components. Given a standardized sketch $\widetilde{S_i}$, the view estimation module first estimates its viewpoint. $\widetilde{S_i}$ is then fed to the sketch-to-3D module to generate a point cloud $P_{i,pre}$, whose pose aligns with the sketch viewpoint. A 3D rotation corresponding to the viewpoint is then applied to $P_{i,pre}$ to output the canonically-posed point cloud $P_i$. The  objective of $G$ is to minimize distances between reconstructed point cloud $P_i$ and the ground-truth point cloud $P_{i,gt}$.

 \textbf{View Estimation Module.} The view estimation module $g_1$ aims to determine the three-dimensional pose from an input sketch $\widetilde{S}$. Similar to the input transformation module of the PointNet \cite{DBLP:journals/corr/QiSMG16}, $g_1$ estimates a 3D rotation matrix $A$ from a sketch $\widetilde{S}$, i.e., $A=g_1(\widetilde{S})$. A regularization loss $L_{\text{orth}}=\|I-AA^T\|^2_F$ is applied to ensure $A$ is a rotation (orthogonal) matrix. The 
rotation matrix $A$ rotates a point cloud from the viewpoint pose to a canonical pose, which matches the ground truth.

 \textbf{3D Reconstruction Module.}
The reconstruction network $g_2$ learns to reconstruct a 3D point cloud $P_{pre}$ from a sketch $\widetilde{S}$, i.e., $P_{pre}=g_2(\widetilde{S})$. $P_{pre}$ is further transformed by the corresponding rotation matrix $A$ to $P$ so that $P$ aligns with the ground-truth 3D point cloud $P_{gt}$'s canonical pose. Overall, we have $P = 
g_1(\widetilde{S})\cdot g_2(\widetilde{S})$. To train $G$, we penalize the distance between an output point cloud $P$ and the ground-truth point cloud $P_{gt}$. We employ the Chamfer distance (CD) between two point clouds $P, P_{gt} \subset \mathbb{R}^3 $:

{\small
\begin{align}
\label{eq:chamfer}
 d_{CD}(P\| P_{gt}) = \sum_{\mathbf{p} \in P} \min_{\mathbf{q} \in P_{gt}} \|\mathbf{p}-\mathbf{q} \|_2^2 + \sum_{\mathbf{q} \in P_{gt}} \min_{ \mathbf{p} \in  P } \|\mathbf{p}-\mathbf{q} \|_2^2 
\end{align}}

The final loss of the entire network is:
\begin{align}
    L & = \sum_i{d_{CD}\left(G\circ U(S_i)\| P_{i,gt}\right) + \lambda L_{\text{orth}}}\\
    &= \sum_i{ d_{CD}\left( A_i \cdot P_{i,pre}\| P_{i,gt}\right) + \lambda L_{\text{orth}}}\\
    &=\sum_i{ d_{CD}\left( g1(\widetilde{S}_i) \cdot g_2(\widetilde{S}_i)\| P_{i,gt}\right) + \lambda} L_{\text{orth}}
\end{align}
where $\lambda$ is the weight of the orthogonal regularization loss and $\widetilde{S_i} = R\circ D_2 \circ D_1(S_i)$ is the standardized sketch from $S_i$. Note that we employ CD rather than EMD (Section \ref{sec:evalm}) to penalize the difference between the reconstruction and the ground-truth point clouds because CD emphasizes the geometric outline of point clouds and leads to reconstructions with better geometric details. EMD, however, emphasizes the point cloud distribution and may not preserve the geometric details well at locations with low point density.

\begin{figure}[t!]
\centering
    \includegraphics[width=0.5\textwidth]{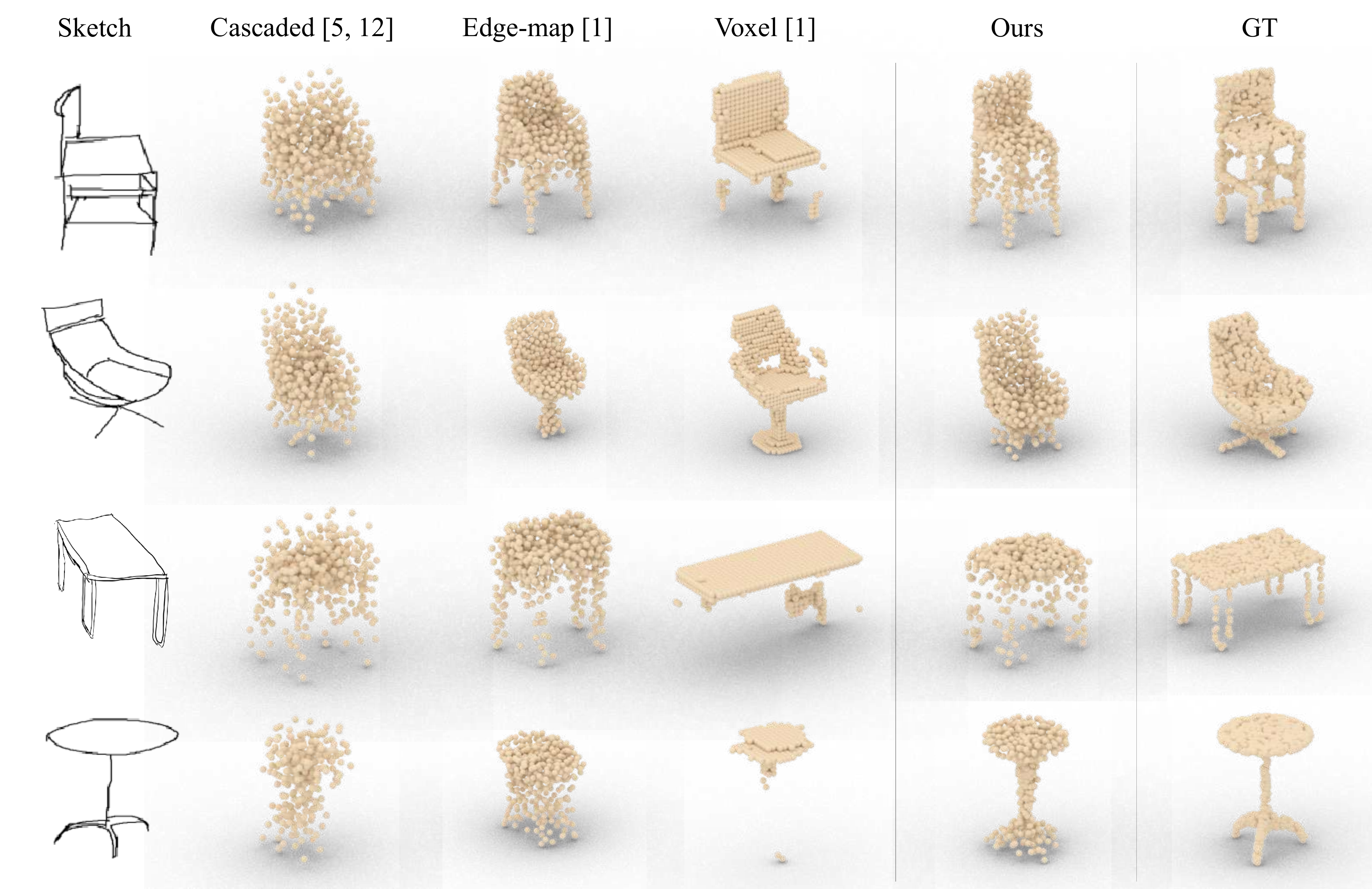}
  \caption{ Performance on free-hand sketches with different design choices. The design pool includes the model with a cascaded two-stage structure (2nd column),  the model trained on edge maps (3rd column),  the model whose 3D output is represented by voxel (4th column), 
  and the proposed model (5th column). Overall, the proposed method achieves better performance and keeps more fine-grained details, e.g., the legs of chairs.}
   \label{fig:comp_real}

\end{figure}

\section{Experimental Results}
\label{sec:exp}
We first present the datasets, training details, evaluation metrics, and implementation details, followed by our model's qualitative and quantitative results. Then, we provide comparisons with some state-of-the-art methods. We also conduct ablation studies to understand the benefits of each module.

{\color{black}
\subsection{3D Sketching Dataset}
\label{sec:dataset}

To evaluate the performance of our method, we collected a real-world evaluation set containing paired sketch-3D data. Specifically, we randomly choose ten 3D shapes from each of the 13 categories of the ShapeNet dataset \cite{shapenet2015}. Then we randomly render 3 images from different viewpoints for each 3D shape. Totally, there are 130 different 3D shapes and 390 rendered images. We recruited 11 volunteers to draw the sketches for the rendered images. Final sketches are reviewed for quality control. We present several examples in Fig.\ref{fig:sketch_gen}.
}

\subsection{ Training Details and Evaluation Metrics}
\label{sec:evalm}
\textbf{Training.} The proposed model is trained on a subset of ShapeNet \cite{shapenet2015} dataset, following settings of \cite{xie2019pix2vox}. The dataset consists of 43,783 3D shapes spanning 13 categories, including car, chair, table, etc. For each category, we randomly select 80\% 3D shapes for training and the rest for evaluation. As mentioned in Section \ref{sec:sk_gen}, corresponding sketches of rendered images from 24 viewpoints of each 3D shape of ShapeNet are synthesized with our synthetic sketch generation module (Section \ref{sec:skgen}). 

\textbf{Evaluation.} To evaluate our method's 3D reconstruction performance on free-hand sketches, we use our proposed sketch-3D datasets (Section \ref{sec:dataset}). To evaluate the generalizability of our model, we also evaluate on three additional free-hand sketch datasets, including the Sketchy dataset \cite{sangkloy2016sketchy}, the TU-Berlin dataset \cite{eitz2012hdhso}, and the QuickDraw dataset \cite{quickdraw}. For these additional datasets, only sketches from categories that overlap with the ShapeNet dataset are considered.

Following the previous works on point cloud generation \cite{DBLP:journals/corr/FanSG16, mandikal20183d, yang2019pointflow}, we adopt two evaluation metrics to measure the similarity between the reconstructed 3D point cloud $P$ and the ground-truth point cloud $P_{gt}$. The first one is the Chamfer Distance (Eqn.\ref{eq:chamfer}), and another one is the Earth Mover’s Distance (EMD):
\begin{align}
    d_{EMD}(P, P_{gt}) = \min_{\phi: P \mapsto P_{gt}} \sum_{x \in P} \| x-\phi(x) \|
\end{align}
where $P, P_{gt} $ has the same size $|P|=|P_{gt}|$ and $\phi: P \mapsto P_{gt}$ is a bijection. CD and EMD evaluate the similarity between two point clouds from two different perspectives (more details can be found in \cite{DBLP:journals/corr/FanSG16}).

\subsection{Implementation Details}

\textbf{Sketch Generation.} We utilize an off-the-shelf sketch-image translation model \cite{liu2019unpaired} to synthesize sketches for training.  
 Given the appropriate quality of the generated sketches on the ShapeNet dataset (with some samples depicted in Fig.\ref{fig:sketch_gen}), we directly use the model without any fine-tuning. 

 \textbf{Data Augmentation.} During training, to further improve the model's generalizability and robustness, we perform data augmentation for synthetic sketches before feeding them to the standardization module. Specifically, we apply image spatial translation (up to $\pm 10$ pixels) and rotation (up to $\pm 10^{\circ}$) on each input sketch. 

 \textbf{Sketch Standardization.} Each input sketch $S_i$ is first randomly deformed with moving least squares \cite{schaefer2006image} both globally and locally ($D_{1}$), and then binarized and dilated five times iteratively ($D_{2}$) to obtain a rough sketch $S_r$. The rough sketch $S_r$ is then used to train a Pix2Pix model \cite{isola2017image}, $R$, to reconstruct the input sketch $S_i$. The network is trained for 100 epochs with an initial learning rate of 2e-4. Adam optimizer \cite{kingma2014adam} is used for the parameter optimization. During evaluation, random deformation $D_{1}$ is discarded.

 \textbf{3D Reconstruction.} The 3D reconstruction network is based on \cite{DBLP:journals/corr/FanSG16}'s framework with hourglass network architecture \cite{newell2016stacked}. We compare several different network architectures (simple encoder-decoder architecture, two-prediction-branch architecture, etc.) and find that hourglass network architecture gives the best performance. This may be due to its ability to extract key points from images \cite{newell2016stacked, cao2017realtime}. We train the network for 260 epochs with an initial learning rate of 3e-5. The weight $\lambda$ of the orthogonal loss is 1e-3. To enhance the performance on every category, all categories of 3D shapes are trained together. The class-aware mini-batch sampling \cite{shen2016relay} is adopted to ensure a balanced category-wise distribution for each mini-batch. We choose Adam optimizer \cite{kingma2014adam} for the parameter optimization. 3D point clouds are visualized with the rendering tool from \cite{mo2019structurenet}.

\begin{table*}[t]
\centering
\LARGE
\caption{ Our approach outperforms baseline methods for 3D shape reconstruction. 
 \cite{delanoy20183d} uses  edge-maps rather than sketches as input.
\cite{xie2019pix2vox} uses voxels rather than point clouds as output. \cite{nozawa2020single} represents the 3D shapewith multi-view depth maps.
``cas.'' refers to the two-stage cascaded training following \cite{liu2019unpaired, DBLP:journals/corr/FanSG16}. CD and EMD measure distances between reconstructions and ground-truths from different perspectives (see text for details). The lower, the better.}

\label{tab:exp_sum}
\resizebox{2\columnwidth}{!}{
\begin{tabular}{c|ccccc>{\color{black}}c|c||ccccc>{\color{black}}c|c}
\hline
\multirow{2}{*}{error} & \multicolumn{7}{c}{\textbf{C}hamfer \textbf{D}istance ($\times 10^{-4}$)}                                                       & \multicolumn{7}{c}{\textbf{E}arth \textbf{M}over's \textbf{D}istance ($\times 10^{-2} $)}                                                          \\ 
\cline{2-15} 
                          & points &  edge \cite{delanoy20183d}  &  voxel \cite{xie2019pix2vox} & cas.  & \cite{nozawa2020single} &retrieval\cite{wang2015sketch} & ours          & points &  edge \cite{delanoy20183d} &  voxel \cite{xie2019pix2vox}       & cas.  & \cite{nozawa2020single} &retrieval\cite{wang2015sketch} & ours          \\ \hline \hline
airplane                  &11.4&7.8&35.1&71.7&8.0&11.2& \textbf{6.1}  &8.5&7.3&10.8&12.7&8.5&11.9& \textbf{6.5}  \\
bench                     &29.2&16.7&202.8&414.1&16.8&14.5& \textbf{13.0} &11.1&8.7&22.0&25.8&10.0&8.6& \textbf{7.8}  \\
cabinet                   &61.7&50.4&59.1&354.5&51.5&45.3&\textbf{39.2}&17.6&17.8&17.0&29.6&18.4&17.2& \textbf{16.0} \\
car                       &20.8&13.3&173.2&114.2&14.1&14.2&\textbf{10.4}&      \textbf{8.9}     &20.0&25.2&20.0&21.6&21.2& 18.0          \\
chair                     &41.8&36.4&108.6&237.1&36.1&33.0& \textbf{26.9} &15.1&15.6&19.4&22.8&16.1&15.3& \textbf{13.0} \\
display                   &68.6&48.3&\textbf{33.1}&340.2&49.3&38.2&37.7&15.5&15.1& \textbf{13.1} &27.9&16.4&14.6& 14.4          \\
lamp                      &63.3&59.4&107.0&214.0&60.2&63.5&\textbf{46.3}&21.3&22.6&21.2&24.9&22.3&22.6& \textbf{20.4} \\
speaker                   &88.2&79.7&203.2&406.4&81.2&72.3&\textbf{62.1}&19.4&19.2&23.8&28.0&21.8& 20.0&\textbf{17.9} \\
rifle                     &17.0&12.1&170.1&15.4&12.3&14.2&\textbf{10.1}&\textbf{11.2}&13.8&23.7&15.4&15.2&17.6& 12.4          \\
sofa                      &32.8&20.9&141.2&482.4&22.3&20.3& \textbf{16.3} &11.1&8.5&18.6&25.4&9.1&8.6& \textbf{7.7}  \\
table                     &55.2&49.4&134.7&469.5&50.5&49.1&\textbf{40.7}&19.1&17.7&18.5&26.5&18.2&18.2& \textbf{17.3} \\
telephone                 &30.7&27.3&26.9&259.8&27.1&27.4&\textbf{21.3}&13.4&13.6&15.1&27.2&15.1&15.3& \textbf{12.3} \\
watercraft                &32.9&26.0&129.1&53.8&26.0&27.3&\textbf{20.3}&12.5&11.1&23.1&17.8&12.2&12.7& \textbf{10.6} \\ \hline
avg.                      &42.6&34.4&117.2&264.1&35.0&33.1&\textbf{26.9}&14.2&14.7&19.3&23.4&15.8&15.7& \textbf{13.4} \\  \hline
free-hand sketch               &87.1&89.0&162.5&334.2&91.8&89.2& \textbf{86.1} &18.6&16.4&22.9&26.1&17.0&16.8& \textbf{16.0} \\ \hline

\end{tabular}}
\end{table*}

\subsection{Results and Comparisons}
\label{sec:edgemap}

We first present our model's 3D shape reconstruction performance, along with the comparisons with various baseline methods.
Then we present the results on sketches from different viewpoints and of different categories, as well the results on other free-hand sketch datasets. Note that unless specifically mentioned, all evaluations are on the free-hand sketches rather than synthesized sketches.

 \textbf{Baseline Methods.} Our 3D reconstruction network is a one-stage model where the input sketch is treated as an image, and point clouds represent the output 3D shape. As conducting the first work for single-view sketch-based 3D reconstruction, we explore different design options adopted by previous works on distortion-free line drawings and/or 3D reconstruction, including architectures, representation of sketches and 3D shapes. We compare with different variants to demonstrate the effectiveness of each choice of our model. 

\textit{Model design: end-to-end vs. two-stage.} Although the task of reconstructing 3D shapes from free-hand sketches is new, sketch-to-image synthesize and 3D shape reconstruction from images have been studied before \cite{liu2019unpaired, xie2019pix2vox, DBLP:journals/corr/FanSG16}. Is a straight combination of the two models, instead of an end-to-end model, enough to perform well for the task? To compare these two architectures' performance, We implement a cascaded model by composing a sketch-to-image model \cite{cyclegan} and an image-to-3D model \cite{DBLP:journals/corr/FanSG16} to reconstruct 3D shapes.

\textit{Sketch: point-based vs. image-based.} 
Considering a sketch is relatively sparse in pixel space and consists of colorless strokes, we can employ 2D point clouds to represent a sketch. Specifically, 512 points are randomly sampled from strokes of each binarized sketch, and we use a point-to-point network architecture (adapted from PointNet \cite{DBLP:journals/corr/QiSMG16}) to reconstruct 3D shapes from the 2D point clouds. 

\textit{Sketch: Using edge maps as proxy.} 
We compare with a previous work \cite{delanoy20183d}. Our proposed model uses synthetic sketch for training. However, an alternative option is using edge maps as a proxy of the free-hand sketch. As edge maps can be generated automatically (we use the Canny edge detector in implementation), the comparison helps us understand if our proposed synthesizing method is necessary. 

\textit{3D shape: voxel vs. point cloud.} We compare with a previous work \cite{xie2019pix2vox}. In this variant, we follow their settings and represent a 3D shape with voxels. As the voxel representation is adopted from the previous method, the comparison helps to understand if representing 3D shapes with point clouds has benefits. 

\textit{3D shape: depth map vs. point cloud.} In this variant, we exactly follow a previous work \cite{nozawa2020single} and represent the 3D shape with multi-view depth maps.

\begin{figure}[t!]
\centering
    \includegraphics[width=0.5\textwidth]{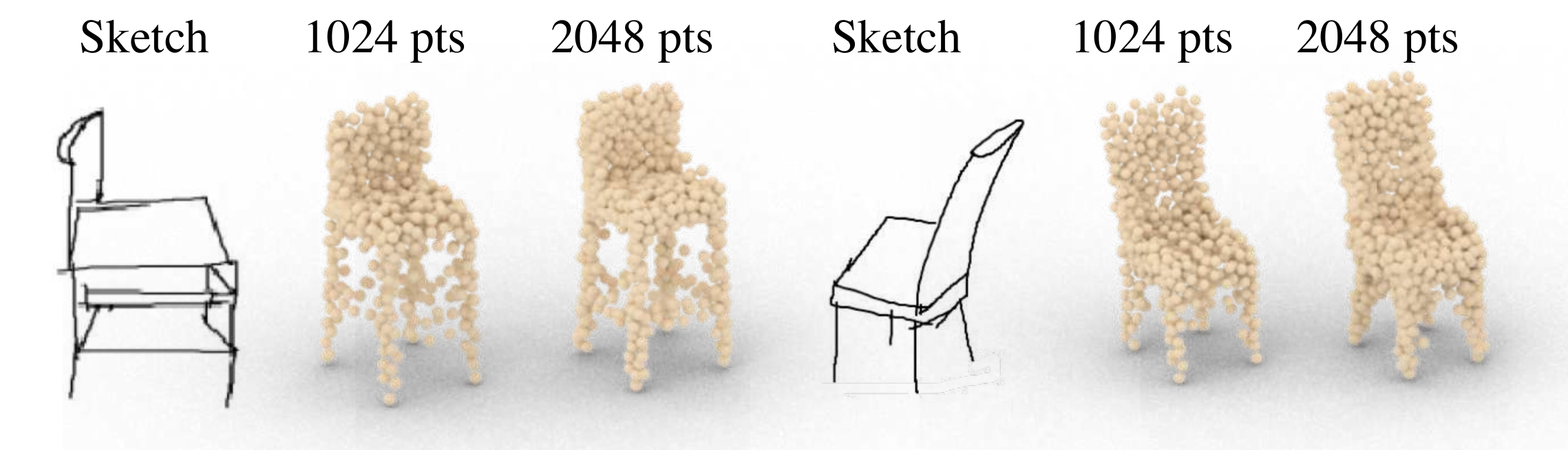}
  \caption{ Reconstruction with different number of points. Increasing the number of points improves the reconstruction quality.}
   \label{fig:numpt}
\end{figure}

\textbf{Comparison and Results.} Table \ref{tab:exp_sum} and Fig.\ref{fig:comp_real} present  quantitative and qualitative results of our method and different design variants. Specifically for quantitative comparisons (Table \ref{tab:exp_sum}), we report 3D shape reconstruction performance on both synthesized (evaluation set) and free-hand sketches. This is due to that the collected free-hand sketch dataset is relatively small and together they provide a more comprehensive evaluation. 
We have the following observations: 
\begin{enumerate}
    \item Representing sketches as images outperforms representing them as 2D point clouds (points vs. ours).
    
    \item  The model trained on synthesized sketches performs better on real free-hand sketches than the model trained on edge maps (89.0 vs. 86.1 on CD, 16.4 vs. 16.0 on EMD). Training with edge maps could reconstruct okay overall coarse shape. However, the unsatisfactory performance on geometric details reveals such methods are hard to generalize to free-hand sketches with distortions. It also shows the necessity of the proposed sketch generation and standardization modules.
    
    \item In terms of the model design, the end-to-end model outperforms the two-stage model by a large margin (cas. vs. ours). 
    
    \item In terms of the 3D shape representation, while the voxel representation can reconstruct the general shape well, the fine-grained details are frequently missing due to its low resolution ($32 \times 32 \times 32$). Thus, point clouds outperform voxels. The proposed method also outperforms a previous work that uses depth maps as 3D shape representation \cite{nozawa2020single} . Note that the resolution of voxels can hardly improve much due to the complexity and computational overhead. 
However, we show that increasing the number of points improves the reconstruction quality (Fig.\ref{fig:numpt}). 
\end{enumerate}

\begin{figure}[t!]
\centering
    \includegraphics[width=0.50\textwidth]{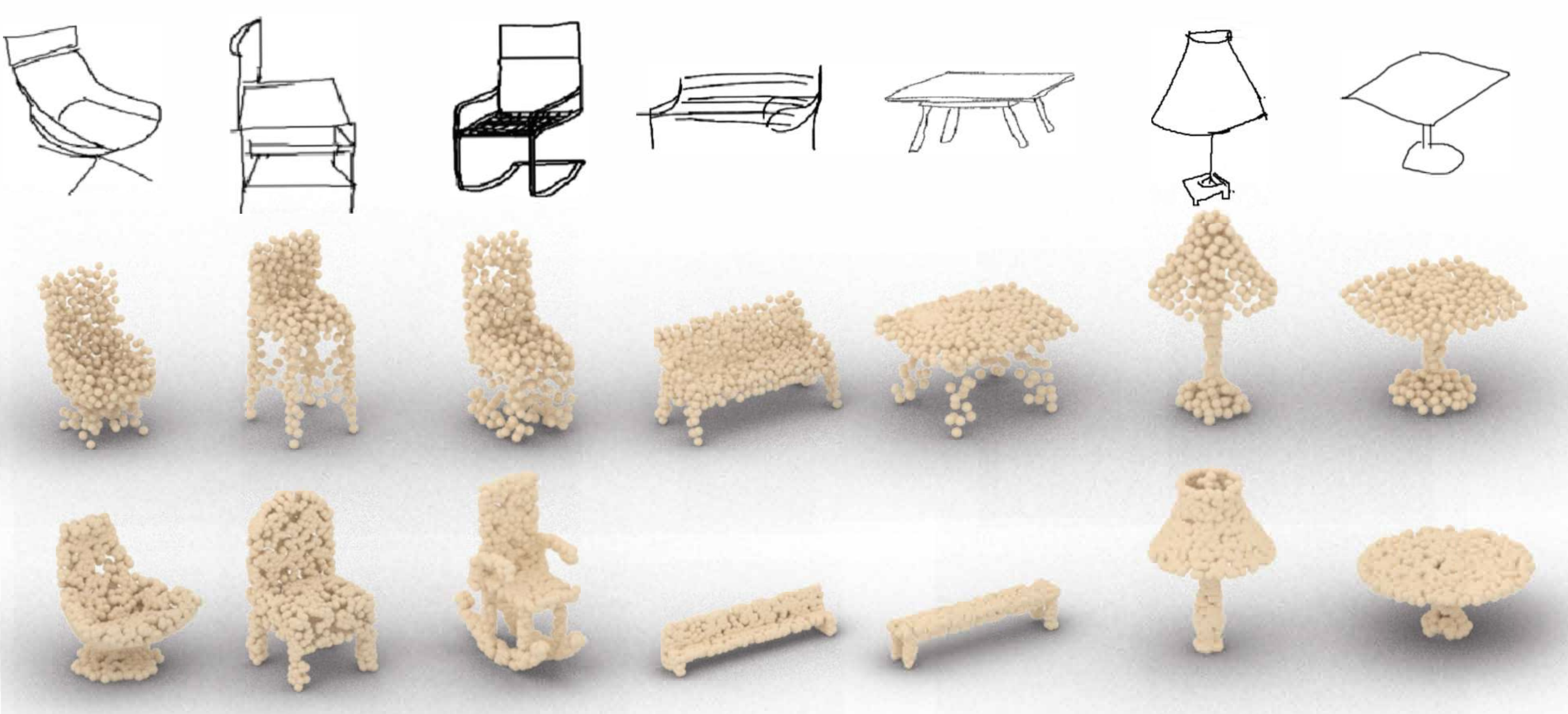}
  \caption{ Ours (2nd row) versus nearest-neighbor retrieval results (last row) of given sketches. Our model generalizes to unseen 3D shapes better and has higher geometry fidelity.}
   \label{fig:retrieval}

\end{figure}

{\color{black} \textbf{Retrieval Results.} We show the generalizability of the proposed method compared with nearest-neighbor retrievals, following methods and settings of \cite{wang2015sketch} (Fig.\ref{fig:retrieval}). Our method is able to generalize to unseen 3D shapes and reconstructs with higher  geometry fidelity (e.g., stand of the lamp, square shape of the table).}

\begin{figure*}[t!]
\centering
  \includegraphics[width=1\textwidth]{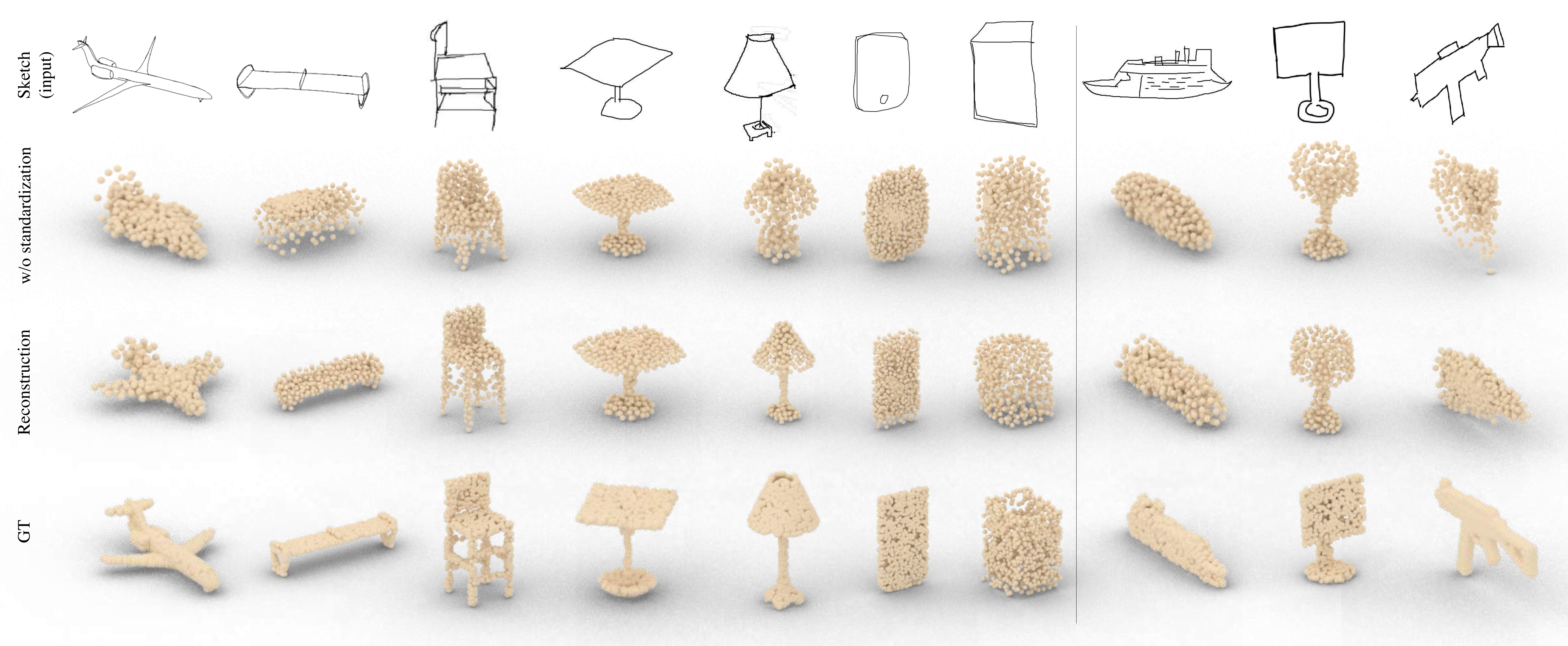}
  \caption{ 3D reconstructions on our newly-collected free-hand sketch evaluation dataset. \textbf{Left:} Examples of some good reconstruction results. Our model reconstructs 3D shapes with fine geometric fidelity of multiple categories \textit{unconditionally}. \textbf{Right:} Examples of failure cases. Our model may not handle detailed structures well (e.g., \textit{watercraft}), recognize the wrong category (e.g., \textit{display} as a \textit{lamp}) due to the ambiguity of the sketch, as well as not able to generate 3D shape from very abstract sketches where few geometric information is available (e.g., rifle). }
  \label{fig:categ}
\end{figure*}

\begin{figure*}[t!]
\centering
  \includegraphics[width=1\textwidth]{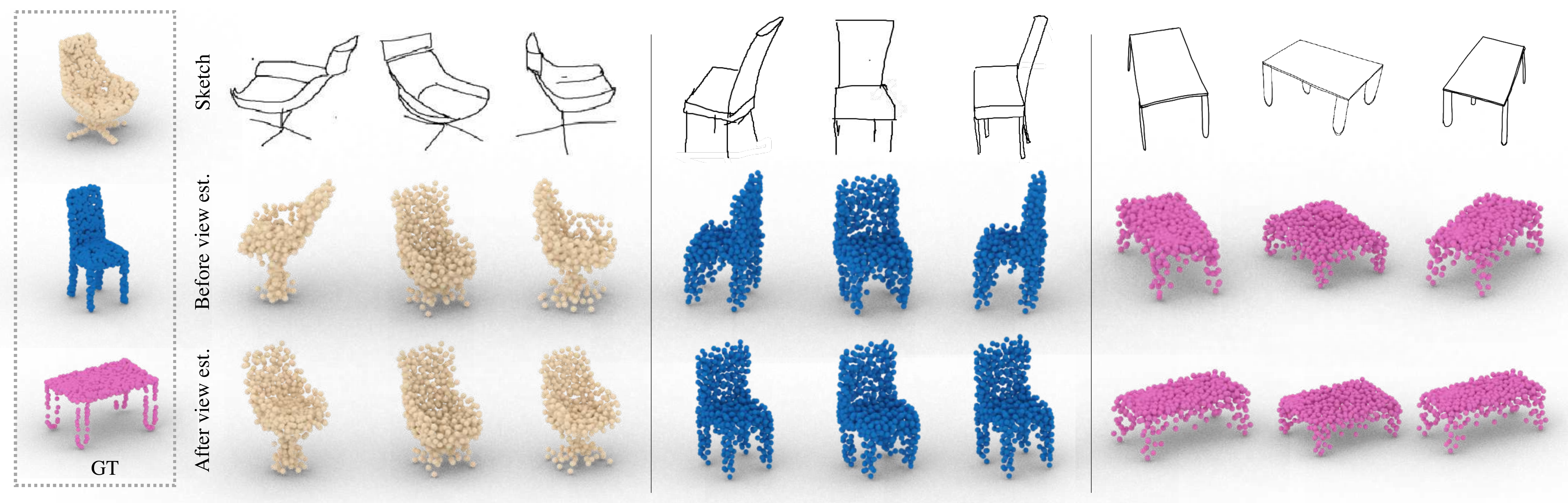}
  \caption{ 3D reconstructions of sketches from different viewpoints. Before the view estimation module, the reconstructed 3D shape aligns with the input sketch's viewpoint. The module transforms the pose of the output 3D shape to align with canonical pose, i.e. the pose of the ground-truth 3D shape. }
  \label{fig:multiview}

\end{figure*}
 \textbf{Reconstruction with Different Categories and Views.}
 Fig.\ref{fig:categ} shows 3D reconstruction results with sketches from different object categories. Our model  reconstructs 3D shapes of multiple categories \textit{unconditionally}. There are some failure cases that the model may not handle well. 

Fig.\ref{fig:multiview} depicts 3D reconstructions with sketches from different views. Our model can reconstruct 3D shapes from different views even if certain parts are occluded (e.g. legs of the table). Slight variations in details exist for different views. 

\begin{figure*}[t!]
\centering
  \includegraphics[width=1\textwidth]{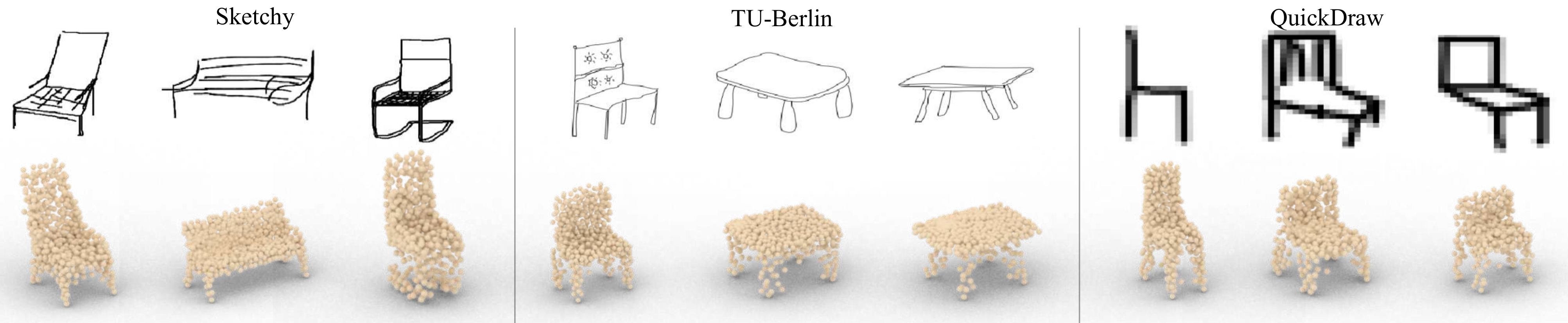}
  \caption{
  Our approach trained on ShapeNet can be directly applied to other unseen sketch datasets and it genealize well. 
  Results on other sketch datasets. \textbf{Left}: Sketchy dataset \cite{sangkloy2016sketchy}; \textbf{Middle}: TU-Berlin dataset \cite{eitz2012hdhso}; \textbf{Right}: QuickDraw dataset \cite{quickdraw}. Our model is able to reconstruct 3D shapes from sketches with different styles and line-widths, and even low-resolution data.}
  \label{fig:sketchy}
\end{figure*}

 \textbf{Evaluation on Other Free-Hand Sketch Datasets.} To evaluate the generalizability of the proposed method, we also evaluate it on three other free-hand sketch datasets \cite{sangkloy2016sketchy, eitz2012hdhso, quickdraw}. We only present some qualitative results (Fig.\ref{fig:sketchy}) as the ground-truth 3D shapes are not available. Our model can reconstruct 3D shapes from sketches with different styles, line-widths, and levels of distortions even at low resolution.

\begin{figure*}[t!]
\centering
  \includegraphics[width=1\textwidth]{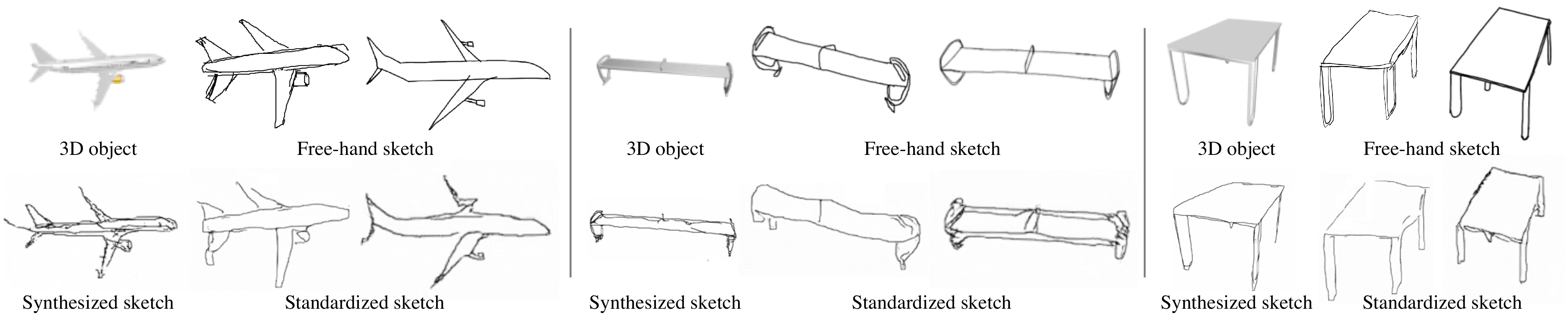}
  \caption{ Standardized sketches converted from different individual styles (by different volunteers). For each rendered image of a 3D object, we show free-hand sketches from two volunteers and the standardized sketches from these free-hand sketches. Contents are preserved after the standardization process, and standardized sketches share the style similar to the synthesized ones.}
  \label{fig:style}
\end{figure*}

\begin{table}[]
\centering
\normalsize
\caption{ 3D shape reconstruction errors of ablation studies of standardization and view estimation module. Having both standardization and view estimation module gives the highest performance. The lower, the better. }
\label{tab:ablation}

\resizebox{0.89\columnwidth}{!}{\begin{tabular}{c|c c|c}
\hline
 error   & no standardization & no view est. & ours \\ \hline\hline
CD ($\times 10^{-4}$)  & 92.6           & 86.8              & \textbf{86.1} \\ \hline
EMD ($\times 10^{-2}$) & 18.2           & 16.2              & \textbf{16.0} \\ \hline
\end{tabular}}
\end{table}

{\color{black}
\subsection{Sketch  Standardization Module}

\label{sec:ssmexp}
\textbf{Visualization.} The standardization module can be considered as a domain translation module designed for sketches. Fig.\ref{fig:style} shows several free-hand sketches of the same objects drawn by different volunteers. We show the standardized sketches of these free-hand sketches and compare them to the synthesized ones. Applied with the standardization module, the sketches share a style similar to synthesized sketches which are used as training data. Thus, the standardization module helps diminish the domain gap of sketches with various styles and enhances the generalizability of the proposed method. 

\textbf{Ablation Studies of the Entire Module.} The sketch standardization module is introduced to handle various drawing styles of humans. We thus verify this module's effectiveness on real sketches, both quantitatively (Table \ref{tab:ablation}) and qualitatively (Fig.~\ref{fig:categ}). As shown in Table \ref{tab:ablation}, the reconstruction performance has a significant drop when removing the standardization module. Its effect is also proved in visualizations. In Fig.~\ref{fig:categ}, we can observe that our full model equipping with the standardization module can produce 3D shapes with higher quality, being more similar to GT shapes, e.g., the airplane and the lamp. 

\textbf{Ablation Studies of Different Components.} The standardization module consists of two components: sketch deformation and style translation. We study each module's performance and report in Table \ref{tab:ablationsk}. We observe that the style transformation part improves the reconstruction performance better compared with the deformation part, while having both parts gives the highest performance.

\begin{table}[]
\centering
\normalsize
\caption{ \textcolor{black}{ Reconstruction error with different components of the standardization module. Having both parts gives the highest performance. }}
\label{tab:ablationsk}

\resizebox{0.9\columnwidth}{!}{\begin{tabular}{cc |c c }\hline
deformation & translation & CD ($\times 10^{-4}$)           & EMD     ($\times 10^{-2}$)       \\ \hline\hline
 $\times$  & $\times$  & 92.6          & 18.2          \\ \hline
 $\times$  & \checkmark  & 87.2          & 16.3          \\ \hline
 \checkmark   & $\times$ & 90.1          & 17.4          \\ \hline
  \checkmark  &\checkmark  & \textbf{86.1} & \textbf{16.0} \\ \hline
\end{tabular}}
\end{table}

\textbf{Additional Applications}. We show the effectiveness of the proposed sketch standardization with two more applications. The first applications is on cross-dataset sketch classification. We identify the common $98$ categories of TU-Berlin sketch dataset \cite{eitz2012hdhso} and Sketchy dataset \cite{sangkloy2016sketchy}. Then we train on TU-Berlin and evaluate on Sketchy. As reported in Table \ref{tab:zeroshot}, adding additional sketch standardization module, the classification accuracy improves $3$ percentage points.

The second application corresponds to the second example depicted in Fig.\ref{fig:zerodemo}. The target domain is CityScapes dataset \cite{Cordts2016Cityscapes}, where the training data comes from. We extract corresponding edge maps with a deep learning approach \cite{poma2020dense} and train an image-to-image translation model \cite{isola2017image} to translate edge maps to the corresponding RGB images. We evaluate the zero-shot domain translation performance on three new datasets: UNDD \cite{nag2019s} (night images), Night-Time Driving \cite{dai2018dark} (night images) and GTA \cite{richter2016playing} (synthetic images; screenshots taken from simulated environment). The novel domains of night and simulated images can be translated to the target domain of daytime and real-world images.  We visualize the results in Fig.\ref{fig:zeroshot}.

\begin{figure}[t!]
\centering
  \includegraphics[width=1\columnwidth]{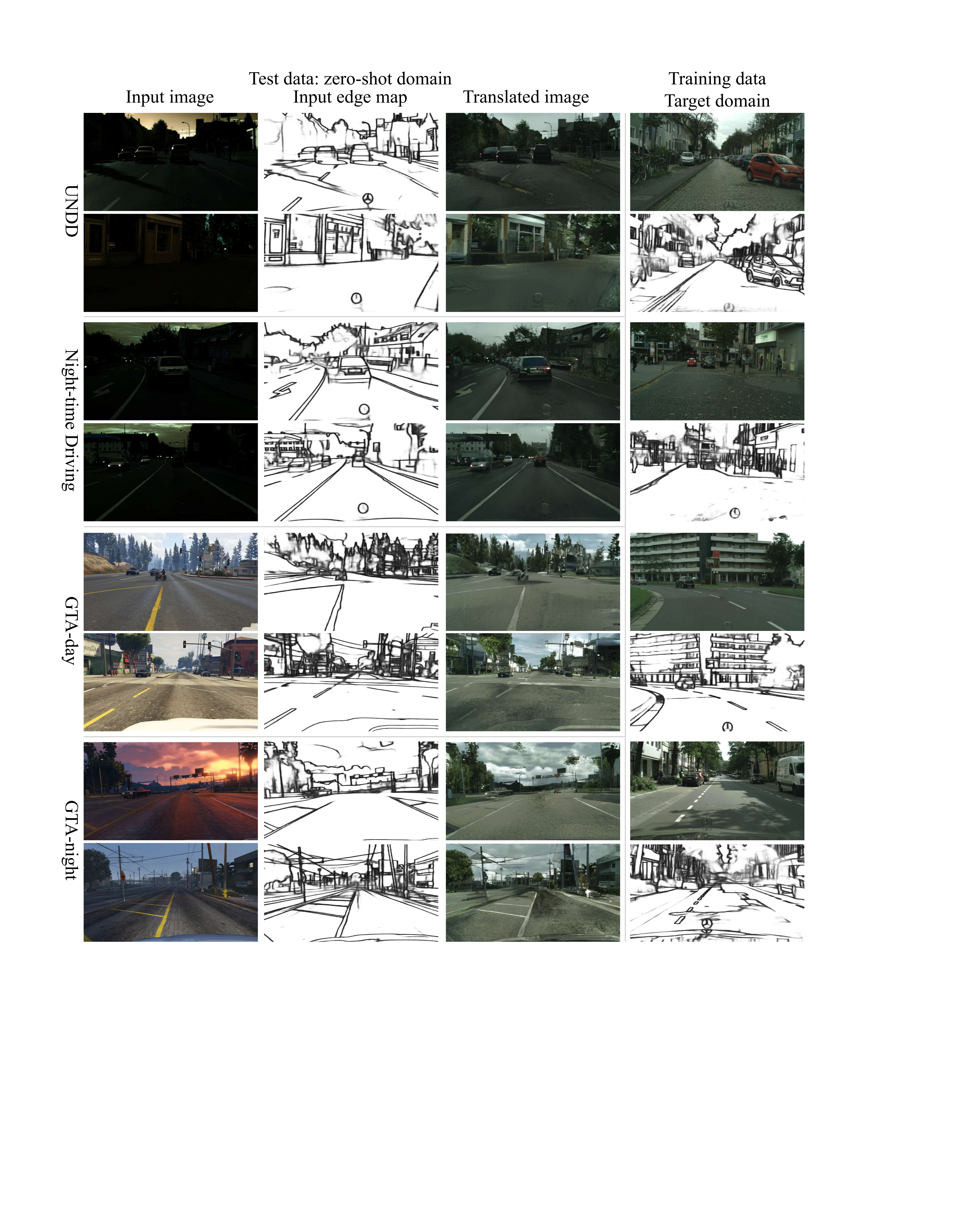}
  \caption{ \textcolor{black}{Zero-shot domain translation results.
  We aim to translate zero-shot images (\textbf{Left}) to the training data domain (\textbf{Right}). Specifically, we evaluate the proposed zero-shot domain translation performance on three new datasets: UNDD \cite{nag2019s}, Night-Time Driving \cite{dai2018dark} and GTA \cite{richter2016playing}. The novel domains of night-time and simulated images can be translated to the target domain of daytime and real-world images by leveraging the synthetic edge map domain as a bridge.  The target domain is CityScapes dataset \cite{Cordts2016Cityscapes}, from where the training data comes. We extract corresponding edge maps and train an image-to-image translation model \cite{isola2017image} to translate edge maps to the corresponding RGB images. 1st, 3rd, 5th, 7th rows of column 1 depict some sample training RGB images and 2nd, 4th, 6th, 8th rows of column 1 depict the corresponding edge maps respectively.
  } }
  \label{fig:zeroshot}
\end{figure}

\begin{table}[]
\centering
\normalsize
\caption{ \textcolor{black}{The sketch standardization module improves cross-dataset sketch classification accuracy. A ResNet-50 model is trained on TU-Berlin \cite{eitz2012hdhso} and evaluated Sketchy dataset \cite{sangkloy2016sketchy}. The sketch standardization module gives 3 percentage points gain.   }}
\label{tab:zeroshot}

\resizebox{0.5\columnwidth}{!}{\begin{tabular}{c|c}
\hline
accuracy  & Sketchy classification                         \\ \hline\hline
w/o std. & 75.1\%                            \\ \hline
w/ std.  & \textbf{78.1\%}                        \\ \hline
\end{tabular}}
\end{table}

}

\subsection{View Estimation Module}

In terms of quantitative results (Table \ref{tab:ablation}), removing the view estimation module leads to a performance drop of CD and EMD. This suggests that an explicit view estimation helps reconstruct the 3D shape more faithfully. The qualitative results of the view estimation module are in Fig.\ref{fig:multiview}. Before the 3D rotation, the reconstructed 3D shape has the pose aligned with the input sketch. After the 3D rotation based on the estimated viewpoint, the 3D shape is aligned to the ground truth's canonical pose.

\section{Summary}
We study a novel task, 3D shape reconstruction from a single-view free-hand sketch.
The major novelty is that we use synthesized sketches as training data and introduce a sketch standardization module, in order to tackle the data insufficiency and sketch style variation issues.   
Extensive experimental results shows that the proposed method is able to successfully reconstruct 3D shapes from single-view free-hand sketches \textit{unconditioned} on viewpoints and categories. 
The work may unleash more potentials of the sketch in applications such as sketch-based 3D design/games, making them more accessible to the general public.

\ifCLASSOPTIONcaptionsoff
  \newpage
\fi

\bibliographystyle{IEEEtran}
\bibliography{ref}

\end{document}


%
\title{3D Shape Reconstruction from Free-Hand Sketches}

%
%
%

\author{Jiayun Wang, Jierui Lin, Qian Yu, Runtao Liu, Yubei Chen, and Stella X. Yu
\thanks{All the authors except Q. Yu are with UC Berkeley / ICSI, Berkeley, CA, USA (e-mail: \{peterwg,jerrylin0928,runtao\_liu,yubeic,stellayu\}@berkeley.edu).}
\thanks{Q. Yu is with Beihang University, Beijing, China (e-mail: qianyu@buaa.edu.cn) and UC Berkeley / ICSI, Berkeley, CA, USA.}
}

%
%

\markboth{IEEE TRANSACTIONS ON IMAGE PROCESSING, submitted}%
{Wang \MakeLowercase{\textit{et al.}}: 3D Shape Reconstruction from Free-Hand Sketches}
%



\maketitle

%
\IEEEpeerreviewmaketitle

\appendices
\input{sections/supp}



\ifCLASSOPTIONcaptionsoff
  \newpage
\fi



%
\bibliographystyle{IEEEtran}
\bibliography{ref}

%



